\documentclass[letter, 10pt, conference]{ieeeconf}  

\usepackage{graphicx}
\usepackage{cite}

\IEEEoverridecommandlockouts                              

\usepackage{epsfig} 

\title{\LARGE \bf Soft Pneumatic Gelatin Actuator for Edible Robotics}

\author{Jun Shintake$^{1}$, Harshal Sonar$^{2}$, Egor Piskarev$^{3}$, Jamie Paik$^{2}$ and Dario Floreano$^{1}$
\thanks{This work was supported by the Swiss National Centre of Competence in Research (NCCR) Robotics.}
\thanks{$^{1}$J.~Shintake and D.~Floreano are with the Laboratory of Intelligent Systems (LIS), \'{E}cole Polytechnique F\'{e}d\'{e}rale de Lausanne, Route Cantonale, Lausanne 1015, Switzerland.
        {\tt\small jun.shintake@epfl.ch}}%
\thanks{$^{2}$H.~Sonar and J.~Paik are with the Reconfigurable Robotics Laboratory (RRL), \'{E}cole Polytechnique F\'{e}d\'{e}rale de Lausanne, Route Cantonale, Lausanne 1015, Switzerland.}
\thanks{$^{3}$E.~Piskarev is with the Section of Mechanical Engineering, \'{E}cole Polytechnique F\'{e}d\'{e}rale de Lausanne, Route Cantonale, Lausanne 1015, Switzerland.}
}

\begin{document}

\maketitle
\thispagestyle{empty}
\pagestyle{empty}

\begin{abstract}
We present a fully edible pneumatic actuator based on gelatin-glycerol composite. The actuator is monolithic, fabricated via a molding process, and measures 90~mm in length, 20~mm in width, and 17~mm in thickness.
Thanks to the composite mechanical characteristics similar to those of silicone elastomers, the actuator exhibits a bending angle of 170.3~$^{\circ}$ and a blocked force of 0.34~N at the applied pressure of 25~kPa.
These values are comparable to elastomer based pneumatic actuators.
As a validation example, two actuators are integrated to form a gripper capable of handling various objects, highlighting the high performance and applicability of the edible actuator. These edible actuators, combined with other recent edible materials and electronics, could lay the foundation for a new type of edible robots.
\end{abstract}

\section{Introduction}
\label{sec:intro}
Revisiting the material foundations of robotic components can lead to novel functionalities and application fields. For example, soft robotics, where robots are composed of compliant materials has led to superior features, such as improved mechanical robustness and simplified structure and control, and safer and more compliant functionalities \cite{soft_robotics, soft_robotics2}.

Here we propose the use of edible materials for a new type of robotic architectures, which we call ``Edible Robotics''. Edible robots can be biodegradable, biocompatible, and environmentally sustainable with none or lower level of toxicity. These functionalities can already be seen in organic electronics made of edible materials \cite{edible_electronics1} to fabricate transistors \cite{edible_FET}, sensors \cite{edible_electronics2}, batteries \cite{edible_electronics3, edible_electronics4, edible_electronics5, edible_electronics6}, electrodes \cite{edible_electronics7}, and capacitors \cite{edible_electronics8}.
The (still missing) availability of edible actuators could pave the way to fully edible robots. The components of such edible robots could also me mixed with nutrient or pharmaceutical components for digestion and metabolization.
Potential applications are disposable robots for exploration (as also mentioned in \cite{gelatin_act, jonathan}), digestible robots for medical purposes in humans and animals, and food transportation where the robot does not require additional payload because the robot is the food.

\begin{figure}[t!]
  \begin{center}
  \includegraphics[width=0.43\textwidth]{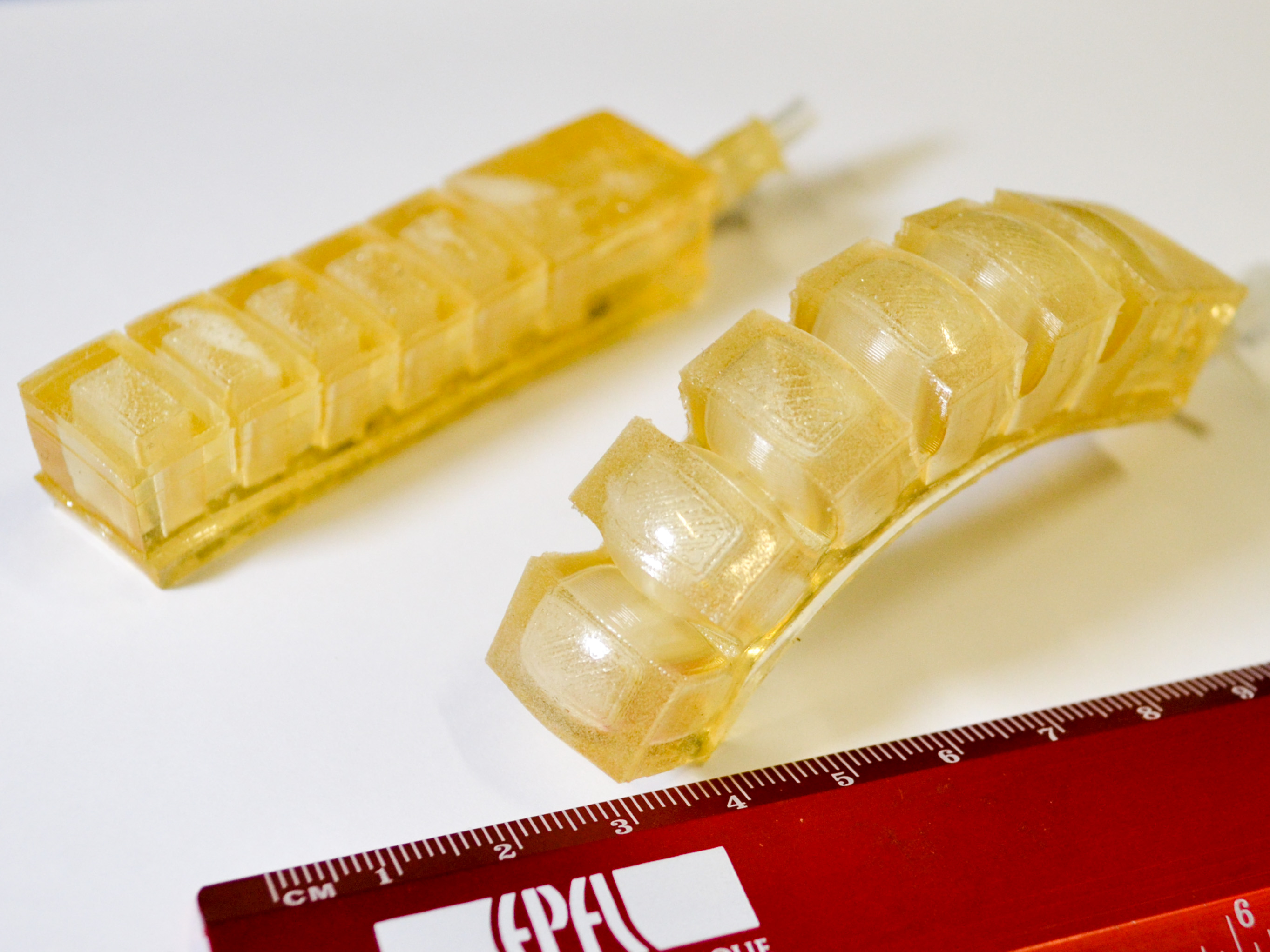}
 \caption{Edible soft pneumatic actuator developed in this study. (left)~actuator in the non-pressurized state. (right)~actuator in a pressurized state.}
 \vspace{-5mm}
    \label{fig:act}
   \end{center}
\end{figure}

So far, the use of edible materials in robotics remains very limited.
Researchers have developed a gelatin hydrogel actuator immersed in NaOH solution \cite{gelatin_act}, and an ingestible robot that uses intestines of pigs as folding parts \cite{pig_foldable}. Both cases require external electric and magnetic fields, which can make them challenging to use outside of a specific environment.

In this paper, we describe a soft gelatin pneumatic actuator for edible robotics. Soft actuators, especially the ones with pneumatic configuration, have being widely developed and applied in various robotic applications thanks to their simple structure that can be extended to form a robot itself \cite{soft_robotics, soft_robotics2}. Therefore, developing edible soft pneumatic actuator is a promising approach that can immediately lead to realization of edible robots. Gelatin is an edible material, and can be polymerized when an edible plasticizer, (glycerol in this work), is used to form a protein network \cite{edible_film}. Gelatin-glycerol composite could replace silicone elastomers, a compliant material often used in soft robots. Once dissolved into aqueous solution, the composite can be used in molding fabrication of soft actuators in the same manner as more traditional elastomers.

In Section II, we perform tensile testing of the gelatin-glycerol composite in order to investigate if its major mechanical properties such as, the Young's modulus, tensile strength, and elongation at break, are comparable to that of silicone elastomers. These results are important for the applicability of the proposed edible material composition for real world robotic applications.
In Section III, we develop a proof-of-concept fully edible soft pneumatic actuator, and characterize its performance in terms of the bending angle and the blocked force as functions of the applied pressure. In Section IV, we show the integration of two edible actuators to form a two-finger gripper capable of grasping various objects. We discuss the results in Section V and draw conclusions in Section VI.



\section{Characterization of gelatin-glycerol membrane}
\label{sec:sample}
In this section, the mechanical properties of the gelatin-glycerol composite is investigated by performing tensile testing.


\subsection{Sample preparation}
Gelatin (48723) and glycerol (G5516) were purchased from Sigma Aldrich (Switzerland).
These materials were used as received.
Aqueous solutions containing gelatin (GEL), glycerol (GLY), and distilled water (WAT) were prepared by mixing the contents for 30~min at 80~$^{\circ}$C.
Two different solutions were prepared: GEL/GLY/WAT = 1:1:8 and 1:2:8.
The solutions were cast on an acrylic mold (20~g solution/mold) and dried at room temperature (25~$^{\circ}$C) for 48~h inside a chemical hood.
After drying, a membrane thickness of 0.5~mm was formed in the mold.
The membrane was then cut by a laser machine (Speedy 300, Trotec Laser) to obtain the dogbone shaped samples (Fig.~\ref{fig:sample}) for uniaxial tensile testing.
This dogbone shape sample was fabricated according to ASTM guidelines \cite{ASTM}.  
The drying time of 48~h was experimentally determined prior to the sample preparation, based on the changes observed for mass of the membrane, which became almost constant at 48~h (Fig.~\ref{fig:eva_stress}(a)).


\subsection{Tensile testing}
Tensile testing of the samples was performed using a mechanical testing machine (5965, Instron).
5 samples were measured for each composition (GEL/GLY = 1:1 and 1:2).
Every sample was pulled uniaxially until it broke (elongation at break), while recording the elongation and the loading. 
In order to obtain Young's modulus and the tensile strength of the gelatin-glycerol composite, the Yeoh hyperelastic material model \cite{Yeoh} was fit to the measured data.

\begin{figure}[t!]
  \begin{center}
  \includegraphics[width=0.45\textwidth]{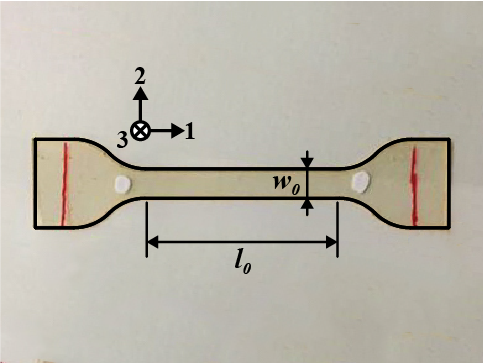}
 \caption{A sample of gelatin-glycerol composite for the tensile testing. Each sample has the same dimensions: length $l_0$ = 35~mm, width $w_0$ = 5~mm, and thickness $h_0$ = 0.5~mm.}
    \label{fig:sample}
   \end{center}
\end{figure}

\begin{figure*}[t!]
  \begin{center}
  \includegraphics[width=0.8\textwidth]{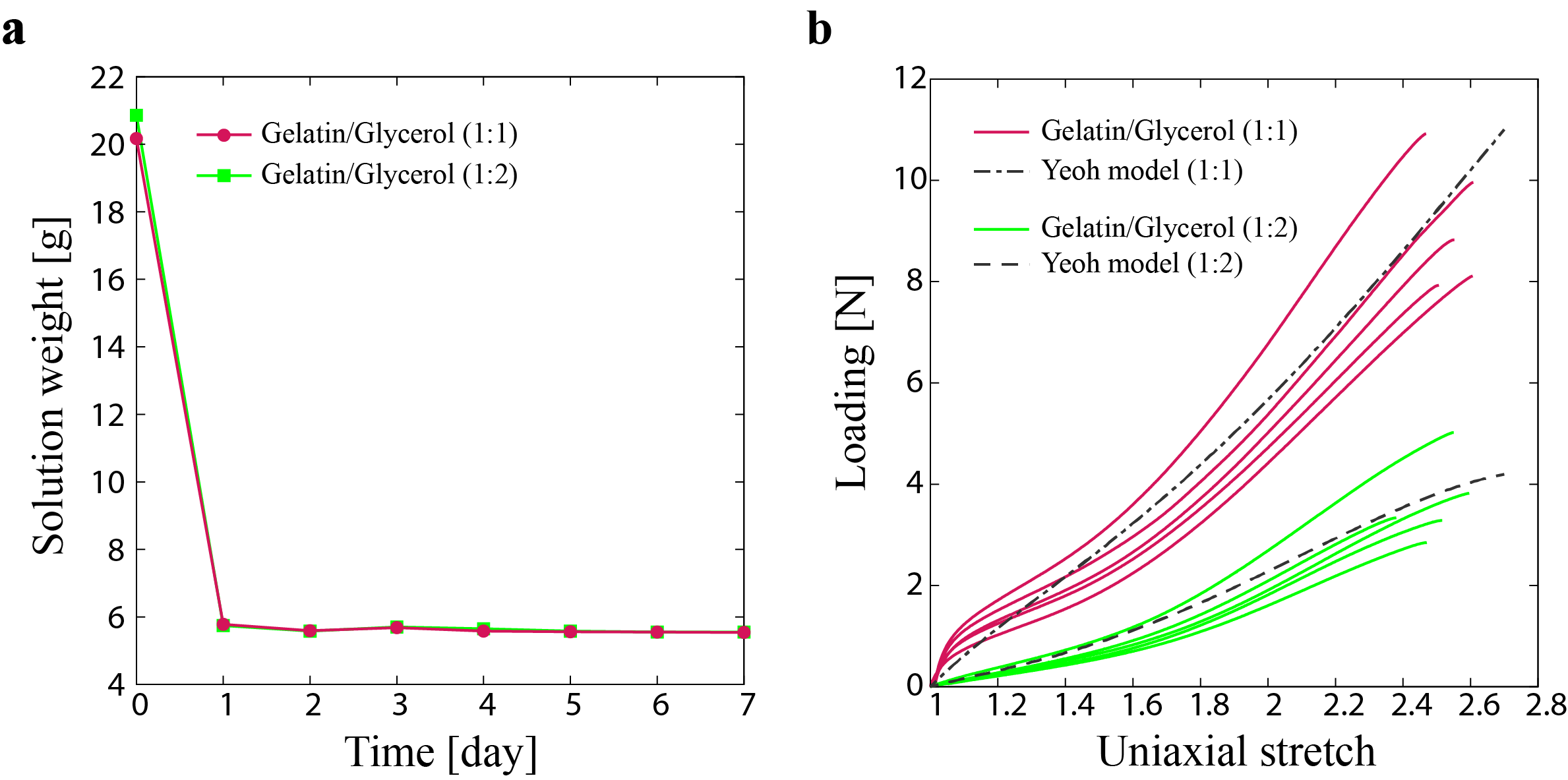}
 \caption{Characterization results for the gelatin/glycerol composite samples. (a)~Variation in the mass of the compositions over time. The water content takes equilibrium after 48~h. (b)~Loading as a function of uniaxial stretch for the two different compositions. Dashed lines represent calculated result using Eq.~\ref{eq:loading}.}
    \label{fig:eva_stress}
   \end{center}
\end{figure*}

The strain energy density function of the Yeoh model is written as

\begin{equation}
W = \sum\limits_{i=1}^3{C_i}(I_1 - 3)^i,
\label{eq:yeoh}
\end{equation}

\noindent where, $C_i$ are material constants. $I_1$ is the strain invariant 

\begin{equation}
I_1 = \lambda_1^2 + \lambda_2^2 + \lambda_3^2,
\label{eq:inv}
\end{equation}

\noindent where, $\lambda_1$, $\lambda_2$, and $\lambda_3$ are the stretch ratio in the length (tensile direction), the width, and the thickness direction, respectively, as represented in Fig.~\ref{fig:sample}. The stretch ratios are defined as

\begin{equation}
\lambda_1 = \frac{l}{l_0}, \lambda_2 = \frac{w}{w_0}, \lambda_3 = \frac{h}{h_0},
\end{equation}

\noindent where, $l$ is the length, $w$ the width, and $h$ is the thickness of the sample.
The subscription 0 stands for their initial state.

We assume the gelatin-glycerol composite to be incompressible, that is

\begin{equation}
\lambda_1\lambda_2\lambda_3 = 1.
\end{equation}

Since the sample is pulled uniaxially, the strains in the width and the thickness directions are the same

\begin{equation}
\lambda_2 = \lambda_3 = \frac{1}{\sqrt{\lambda_1}}.
\label{eq:lambda23}
\end{equation}

By substituting Eq.~\ref{eq:lambda23} and Eq.~\ref{eq:inv} into Eq.~\ref{eq:yeoh}, we get an expression of the strain energy density as a function of $\lambda_1$

\begin{equation}
W = \sum\limits_{i=1}^3{C_i}(\lambda_1^2 + \frac{2}{\lambda_1} - 3)^i.
\end{equation}

With the consideration of the Cauchy stress for incompressible Yeoh hyperelastic material model and uniaxial stretch, the stress in the pulling direction $\sigma_1$ is given by

\begin{equation}
\sigma_1 = {\lambda_1}\frac{\partial W}{\partial \lambda_1} = 2(\lambda_1^2 - \frac{1}{\lambda_1})\sum\limits_{i=1}^3iC_i(\lambda_1^2 + \frac{2}{\lambda_1} - 3)^{i-1}.
\end{equation}

The uniaxial loading $F_1$ is then expressed as

\begin{equation}
F_1 = 2{h_0}{w_0}(\lambda_1 - \frac{1}{\lambda_1^2})\sum\limits_{i=1}^3iC_i(\lambda_1^2 + \frac{2}{\lambda_1} - 3)^{i-1}.
\label{eq:loading}
\end{equation}

Once the loading is measured, the tensile strength $\sigma_{\rm TS}$ at elongation at break $\lambda_{\rm EAB}$ is given as

\begin{equation}
\sigma_{\rm TS} = \frac{F_{1\rm m}}{w_0h_0}\lambda_{\rm EAB},
\end{equation}

\noindent where $F_{1\rm m}$ is the measured uniaxial loading at $\lambda_{\rm EAB}$.
Fitting Eq.~\ref{eq:loading} to the measured loading data gives us the material constants $C_1$, $C_2$, and $C_3$.
The material constants are then used to obtain Young's modulus based on the consistency condition for the Yeoh hyperelastic material model

\begin{equation}
2C_1 = \mu,
\end{equation}

\noindent where $\mu$ is the shear modulus.
We assume the Poisson's ratio of the gelatin-glycerol composite $\nu$ can be approximated to 0.5, which is the case of silicone elastomers.
The Young's modulus of the composite $E$ is then given as

\begin{equation}
E = 2\mu(1+\nu).
\end{equation}

\subsection{Results and discussion}
Fig.~\ref{fig:eva_stress}(b) plots the result of the tensile testing for two different gelatin(GEL)/glycerol(GLY) composition of 1:1 and 1:2.
The samples exhibit hyperelastic (nonlinear) behavior similar to silicone elastomers.
The dashed lines shown in the figure are the calculated loading based on Eq.~\ref{eq:loading}.
The material constants and properties obtained are summarized in Table~\ref{tab:mat1} and Table~\ref{tab:mat2}, respectively.
The measured Young's modulus is 2.7~$\pm$~0.5~MPa for GEL/GLY = 1:1, and 0.7~$\pm$~0.2~MPa for GEL/GLY = 1:2, respectively.
These values of the modulus are in the same range of silicone elastomers used for soft robots (0.125~MPa for Smooth-On Ecoflex 00-30 \cite{ecoflex}, and 1.3-3.0~MPa for Dow Corning Sylgard 184 \cite{DC184}).
Also, it can be seen that the composition ratio changes the modulus; larger the glycerol contents, softer the material is.
Therefore, desired modulus may be achieved by adjusting the ratio of gelatin and glycerol.

Similar to the Young's modulus, the tensile strength for the two compositions (9.3~$\pm$~1.2~MPa for GEL/GLY = 1:1, and 3.7~$\pm$~0.9~MPa for GEL/GLY = 1:2) are  in the same range of silicone elastomers (3.5-7.7~MPa for Sylgard 184 \cite{DC184}, and 1.4~MPa for Ecoflex 00-30 \cite{ecoflex_data}).

The elongation at break obtained for the two is almost the same value; 154.8~$\pm$~6.2~\% for GEL/GLY = 1:1, and 150.3~$\pm$~8.2~\% for GEL/GLY = 1:2, respectively.
The value is reasonable for GEL/GLY = 1:1 because silicone elastomers with high modulus exhibit similar elongation at break (80-170~\% for Sylgard 184 \cite{DC184}).
On the other hand, the value of GEL/GLY = 1:2, 150.3~$\pm$~8.2~\% is significantly smaller than those of soft elastomers (900~\% for Ecoflex 00-30 \cite{ecoflex_data}).
This may have resulted from the rough edge of the samples caused by heating during the laser cutting.
The rough-edged sample would have tearing before it arrives the true elongation at break of the material.
A potential solution for this issue is to prepare the samples using only molding process.

\section{Development and characterization of soft pneumatic gelatin actuator}
Based on the results obtained in the previous section, we developed a proof-of-concept fully edible soft pneumatic actuator using the gelatin-glycerol composite. 

\begin{table}[t!]
\caption{Material constants of gelatin(GEL)-glycerol(GLY) composite obtained from tensile testing}
\vspace{-5mm}
\label{tab:mat1}
\begin{center}
  \begin{tabular}{ c c c }
    Constant [MPa] & GEL/GLY (1:1) & GEL/GLY (1:2) \\ \hline
$C_1$ & 0.45~$\pm$~0.09 & 0.12~$\pm$~0.03 \\
  $C_2$ & 5.72~$\pm~1.08\times10^{-2}$ & 4.60~$\pm~0.87\times10^{-2}$ \\
  $C_3$ & -0.21~$\pm~0.11\times10^{-2}$ & -0.33~$\pm~0.07\times10^{-2}$ \\
  \hline
  \end{tabular}
  \end{center}
\end{table}

\begin{table}[t!]
\caption{Material properties of gelatin(GEL)-glycerol(GLY) composite obtained from tensile testing}
\vspace{-5mm}
\label{tab:mat2}
\begin{center}
  \begin{tabular}{ c c c }
    Property & GEL/GLY (1:1) & GEL/GLY (1:2) \\ \hline
Young's modulus [MPa] & 2.7~$\pm$~0.5 & 0.7~$\pm$~0.2\\
  Tensile strength [MPa] & 9.3~$\pm$~1.2 & 3.7~$\pm$~0.9\\
  Elongation at break [\%] & 154.8~$\pm$~6.2 & 150.3~$\pm$~8.2\\
\hline
  \end{tabular}
  \end{center}
\end{table}

\begin{figure*}[t!]
  \begin{center}
  \includegraphics[width=0.875\textwidth]{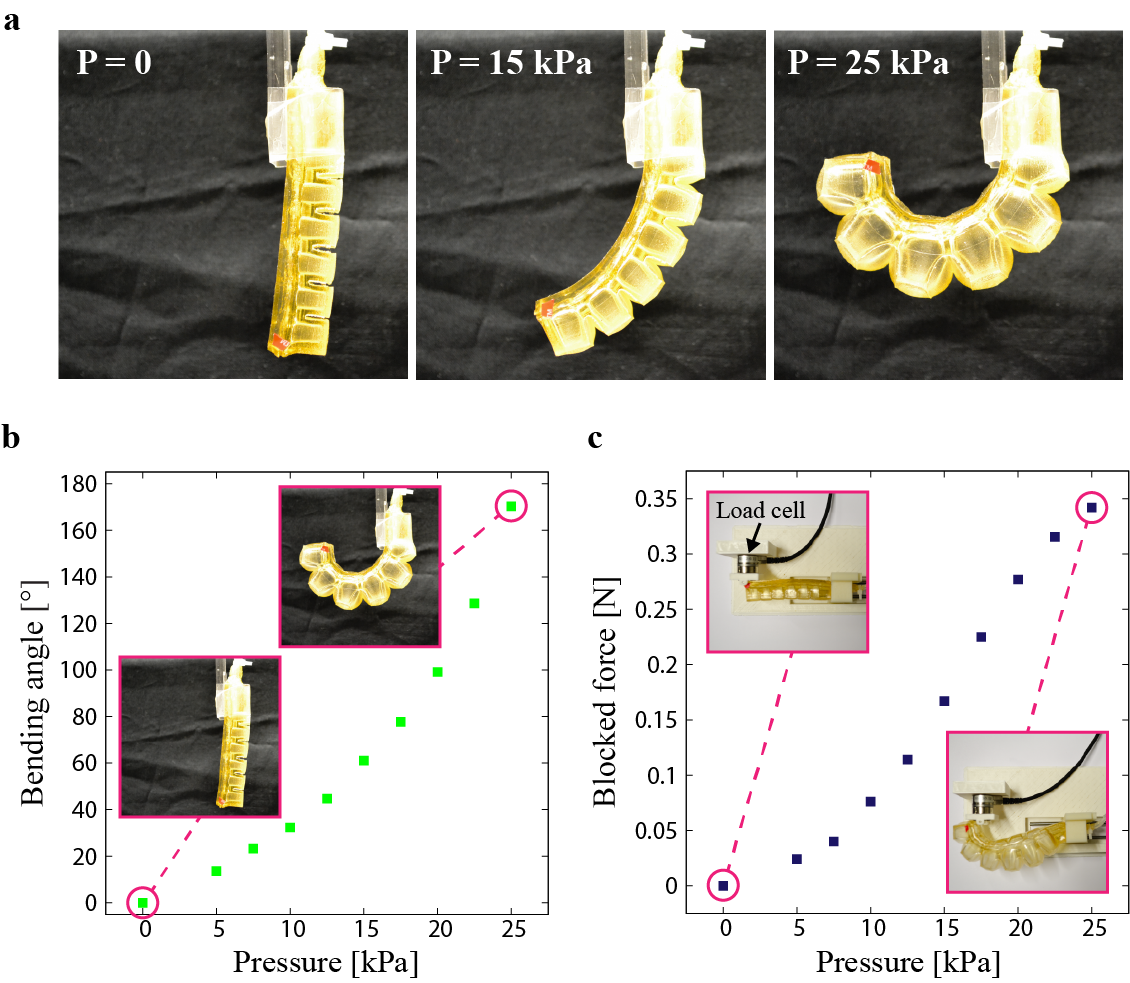}
 \caption{Characterization results of the soft pneumatic gelatin actuator. (a)~Bending of the actuator in different pressured states. P stands for the gauge pressure. (b)~Bending angle as a function of the applied pressure. A bending angle of 170.3~$^{\circ}$ was observed at 25~kPa. (b)~Blocked force as a function of the applied pressure. A force of 0.34~N was measured at 25~kPa.}
    \label{fig:act_test}
   \end{center}
\end{figure*}

\subsection{Design and fabrication}
The actuator shown in Fig.~\ref{fig:act} has multiple, half-separated chambers in the monolithic structure, similar to the configuration initially reported in \cite{multiple_chamber}.
As shown in Fig.~\ref{fig:act_test}(a), when pressurized air is injected, the inflation of the chambers happens in the upper part (shifted position from the neutral plane), resulting in a bending of the structure.
The dimensions of the device were 90~mm in length, 20~mm in width, and 17~mm in thickness.
The actuator was fabricated based on a molding method commonly used for silicone elastomer based pneumatic actuators.

\subsection{Bending angle and blocked force}
The bending angle of the actuator was defined as the tip angle difference from the initial angle, which was recorded by a CMOS camera.
The blocked force of the actuator was measured by putting a load cell (Nano 17, ATI Industrial Automation) on the tip in the way blocking the bending actuation (Fig.~\ref{fig:act_test}(c) inset)
In these characterizations, pressurized air of up to 25~kPa was applied via a setup consisted of a valve, a compressor, and a computer running LabView.

\begin{table}[t!]
\caption{Comparison of blocked force for the edible actuator and other actuators}
\vspace{-5mm}
\label{tab:act_comp}
\begin{center}
  \begin{tabular}{ c c }
Pneumatic actuator & Blocked force [N]\\ \hline
Our edible actuator & 0.3 at 25~kPa\\
Elastomer based actuator in \cite{pneumatic_1} & 0.2 at 25~kPa\\
Elastomer based actuator in \cite{pneumatic_2} & 0.4 at 25~kPa\\
Elastomer based actuator in \cite{pneumatic_3} & 0.5 at 40~kPa\\
Elastomer based actuator in \cite{pneumatic_4} & 0.5 at 50~kPa\\
\hline
  \end{tabular}
  \end{center}
\end{table}

\begin{figure*}[t!]
  \begin{center}
  \includegraphics[width=0.9\textwidth]{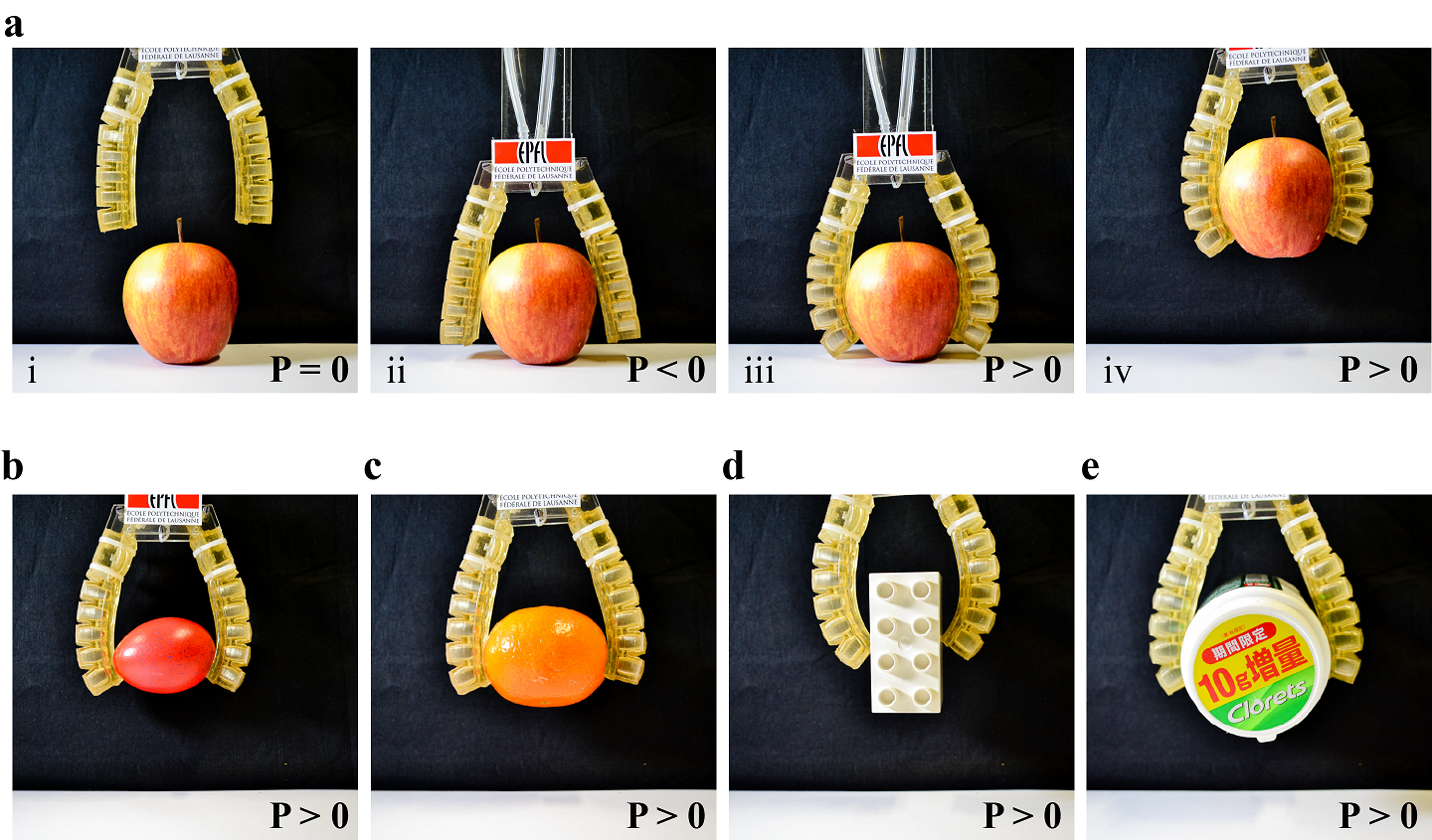}
 \caption{Gripper grasping test. P stands for the gauge pressure. (a-i) The gripper is placed above the object, here an apple mass of 95.6 g. (a-ii) The
device approaches the object. (a-iii) Applying pressure conforms the adaptation of the fingers. (a-iv) The actuated force of the gripper makes it possible to
picked up the object. The gripper demonstrated handling of other objects: (b) a boiled egg (47.7 g), (c) an orange (104.8 g), (e) A LEGO brick (25.7 g),
and (f) A bottle of chewing gums (153.1 g).}
    \label{fig:gripper}
   \end{center}
\end{figure*}

\subsection{Results and discussion}
Fig.~\ref{fig:act_test}(b) plots the bending angle as a function of the applied pressure.
The angle increased with the pressure, and took a value 170.3~$^{\circ}$ at 25~kPa.
Similar trend was observed in the blocked force, and a force value of 0.34~N was recorded at 25~kPa.
These results represent the fact that the performance of the actuator, the actuation angle and the force, are pressure-controllable.
We found our edible soft actuator exhibited comparable performance to existing silicone elastomer based actuators, as summarized in Table~\ref{tab:act_comp}.
During the test, we observed the actuator showed fast motion ($\sim$0.5~s), and was capable of withstand numerous cycles of actuation (see supplemental video).

\section{Demonstration of robotic application}
In order to demonstrate the applicability of the edible
soft pneumatic actuator to robotic application, we integrated
the two actuators into a form of two-finger gripper and
performed grasping test for various objects. The gripper,
shown in Fig.~\ref{fig:gripper}(a-i), has the open-finger at the initial state.
The fingers can further be opened by sucking air from the
chambers (Fig.~\ref{fig:gripper}(a-ii)). When pressurized air is injected,
the fingers adapt to the object, here an apple (95.6 g), and
provide enough force to lift it up (Fig.\ref{fig:gripper}(a-iii-iv)). As shown
in Fig. 3(b-f), we observed the gripper was able to grasp
and lift up varied objects: a boiled egg (47.7 g), an orange
(104.8 g), a LEGO brick (25.7 g), and a bottle of chewing
gums (153.1 g). This versatility of the gripper illustrates high
performance of our edible actuator.

\section{Discussion}
The edible material used for the actuator, the gelatin-glycerol composite, shows similar mechanical properties to those of silicone elastomers.
Assessment of other characteristics such as durability and viscoelasticity with varied composition ratios will provide more insight about applicability of the composite material for robotic applications.
Environmental conditions such as humidity and temperature, are also important aspects of robots that operate in uncontrolled environments.
As discussed in Section II, the gelatin-glycerol composite appears to reach an equilibrium water content (Fig.~\ref{fig:eva_stress}(a)), suggesting that the mechanical properties may change corresponding to humidity of the surrounding air.
However, this is not a major problem because edible coatings to preserve the water content are already available \cite{protection}.
Temperature may also effect to the mechanical properties because the melting temperature of gelatin is around 35~$^{\circ}$C.
The melting point can be higher once plasticizers and other proteins are added. For example, a composite of gelatin, chitosan (a polysaccharide), and glycerol has a melting point of 67~$^{\circ}$C \cite{melting}.
Investigation of above mentioned characteristics will contribute towards the improved design of the actuators and robots, and their applicable tasks.

The soft pneumatic edible actuator described here displays performance comparable to actuators made of silicone elastomer based even if the actuator design described here was not yet optimized. Higher performance may be expected if the actuator is designed with aid of analytical modeling \cite{vito} or FEM \cite{pneumatic_3} where the material properties obtained from the characterization are incorporated. This will also enable the design of actuators of different geometries and sizes. For the future generation of the edible actuators, further characterization of cycles and repeatability should be performed. Given the melting feature of gelatin, our actuator could be capable of self-healing and become re-usable, which are added capabilities that existing soft pneumatic actuators usually do not exhibit. The edible robotic gripper described here shows possibility of creating edible robots based on these materials. Given the simplicity of the actuator design, it could be implemented to many different types of robots.
Along with all the functionalities---biodegradability, biocompatibility, environmental sustainability, digestibility, metabolizability, selfheal ability, and re-usability---those edible robots could bring novel applications.

For example, as discussed in literature \cite{animal} about the animal navigation in wild, fully edible robots would help to study how wild animals collectively behave.
The robots could also take a role of animals prey to observe their hunting behaviors, or to train protected animals to do predation.
Once medical components are mixed into the edible composition, the robots could help preservation of wild animals or heal inside of the human body. 
When edible robots can be metabolized, they also function as energy storage providing an advantage in terms of increased payload with respect to non-edible robots that must be loaded with a food payload.
This would be effective in rescue scenarios where the metabolizable robots can reach survivors in isolated places like inside a crevice or up on mountain.
Last, but not least, since edible materials can generate electric energy \cite{edible_electronics3, edible_electronics4, edible_electronics5, edible_electronics6}, one could envisage autophagy (self-eating) function, like that of octopus \cite{oct}, to extend their lifetime.

\section{Conclusion}
In this paper, we proposed and characterized a gelatin-glycerol composite as a candidate for soft edible actuator material. The characterization results showed that the composite had mechanical properties that were mostly in the same range as those of the conventional silicone elastomers. Based on the results, we also developed a proof of concept, fully edible, soft gelatin pneumatic actuator. The edible actuator exhibited comparable performance to silicone elastomer based actuators. The two actuators were then integrated to form a two-finger gripper. With this gripper, we demonstrated handling of various objects. We believe that edible actuators, combined with other edible electronics and sensors, could pave the way towards the new field of edible robotics.


\section*{Acknowledgments}
This work was supported by the Swiss National Centre of Competence in Research (NCCR) Robotics.

\end{document}